\documentclass[letterpaper, 10 pt, conference]{ieeeconf} 
\IEEEoverridecommandlockouts                              

\usepackage{graphics} 
\usepackage{epsfig} 
\usepackage{times} 
\usepackage{amsmath} 
\usepackage{amssymb}  
\usepackage{color}

\usepackage{comment}
\usepackage{algorithmic, algorithm} 
\usepackage{fourier} 
\usepackage{dblfloatfix} 

\def\mf{\mathbf}

\def\beq{\begin{equation*}}
\def\eeq{\end{equation*}}
\def\bql{\begin{equation}}
\def\eql{\end{equation}}
\def\bqn{\begin{eqnarray*}}
\def\eqn{\end{eqnarray*}}
\def\bnl{\begin{eqnarray}}
\def\enl{\end{eqnarray}}
\def\bna{\bql\begin{array}{rcl}}
\def\ena{\end{array}\eql}
\def\bnn{\beq\begin{array}{rcl}}
\def\enn{\end{array}\eeq}
\def\bma{\begin{bmatrix}}
\def\ema{\end{bmatrix}}
\def\bmx{\begin{matrix}}
\def\emx{\end{matrix}}
\def\ben{\begin{enumerate}}
\def\een{\end{enumerate}}
\def\bit{\begin{itemize}}
\def\eit{\end{itemize}}
\def\bei{\begin{itemize}}
\def\eei{\end{itemize}}
\def\bet{\begin{tabular}}
\def\eet{\end{tabular}}

\newcommand{\allcaps}[1]{\uppercase\expandafter{#1}}

\def\bfs{\begin{footnotesize}}
\def\efs{\end{footnotesize}}
\def\bss{\begin{small}}
\def\ess{\end{small}}

\def\ta{\tau_A(\za,\Gamma)}
\def\td{\tau_D(\zd,\Omega)}
\def\taa{\tau_A(\za,\zb)}
\def\tdd{\tau_D(\zd,\zb)}

\def\zd{\mf z_D}
\def\za{\mf z_A}
\def\zb{\mf z_B}

\def\pp{p(\zd,\za,\zb)}

\def\pdstar{p(\zd,\za,\Omega^*,\Gamma)}
\def\pastar{p(\zd,\za,\Omega,\Gamma^*)}
\def\pdastar{p(\zd,\za,\Omega^*,\Gamma^*)}

\title{\LARGE \bf
Defending a Perimeter from a Ground Intruder Using an Aerial Defender: Theory and Practice
}

\author{Elijah S. Lee$^{1}$, Daigo Shishika$^{2}$, Giuseppe Loianno$^{3}$, and Vijay Kumar$^{1}$

\thanks{We gratefully acknowledge the support from ARL Grant DCIST CRA W911NF-17-2-0181, NSF Grant CNS-1521617,
ARO Grant W911NF-13-1-0350, ONR Grants N00014-20-1-2822 and
ONR grant N00014-20-S-B001, and Qualcomm Research.
}
\thanks{$^{1}$The authors are with the GRASP Lab, University of Pennsylvania, Philadelphia, PA 19104, USA. 
{\tt\footnotesize \{elslee, kumar\}@seas.upenn.edu}
}%
\thanks{$^{2}$The author is with George Mason University, Fairfax, VA 22030, USA. {\tt\footnotesize email: daigo.shishika@gmail.com}
}%
\thanks{$^{3}$ The author is with the New York University, Tandon School of Engineering, Brooklyn, NY 11201, USA. {\tt\footnotesize email: {loiannog}@nyu.edu.}
}%
}

\begin{document}

\maketitle
\thispagestyle{empty}
\pagestyle{empty}

\begin{abstract}
The perimeter defense game has received interest in recent years as a variant of the pursuit-evasion game. A number of previous works have solved this game to obtain the optimal strategies for defender and intruder, but the derived theory considers the players as point particles with first-order assumptions. In this work, we aim to apply the theory derived from the perimeter defense problem to robots with realistic models of actuation and sensing and observe performance discrepancy in relaxing the first-order assumptions. In particular, we focus on the hemisphere perimeter defense problem where a ground intruder tries to reach the base of a hemisphere while an aerial defender constrained to move on the hemisphere aims to capture the intruder. The transition from theory to practice is detailed, and the designed system is simulated in Gazebo. Two metrics for parametric analysis and comparative study are proposed to evaluate the performance discrepancy.   

\end{abstract}

\section{Introduction}
The pursuit-evasion games (PEGs) have been widely investigated over the past years and are used in many applications including mobile robotics \cite{chung2011search}. There are many variants of PEGs under different assumptions on the players and the environments. One route is to consider the players as point particles \cite{liang2019differential, yan2019construction}. Liang et al. \cite{liang2019differential} address a PEG with three point particles: target, attacker, and defender. The attacker aims to capture the target while avoiding the defender, and the defender aims to defend the target while trying to capture the attacker. Other work \cite{yan2019construction} allows the point particle players move in three dimensions and solves the differential games with three defenders and one intruder with equal speeds.

Researchers also have focused on solving the PEGs with real robots \cite{vidal2002probabilistic, alexopoulos2015cooperative, arola2019uav}. Vidal et al. \cite{vidal2002probabilistic} propose a hierarchical hybrid system to implement the pursuit-evasion game scenario on real UAVs and UGVs. The PEG between two pursuing and one evading unmanned aerial vehicles is solved in \cite{alexopoulos2015cooperative}. Deep learning based approach for vision-based UAV pursuit-evasion is implemented in \cite{arola2019uav}.

This work formulates a variant of pursuit-evasion game known as the target-guarding problem \cite{isaacs1999differential}. In this problem, intruders aim to reach the target without being captured by defenders, while defenders try to capture intruders \cite{liang2019differential, shishika2018local, lee2020perimeter-defense}. When defenders are constrained to move along the perimeter of the target region, we call this problem as perimeter defense game, which we refer to \cite{shishika2020review} for detailed survey. Previous works consider the players as point particles with first-order assumptions \cite{liang2019differential,isaacs1999differential,shishika2018local,lee2020perimeter-defense,shishika2020review}. The optimal strategies are theoretically proved for multiplayer game on two-dimensional convex shapes \cite{shishika2018local} and for three-dimensional game on hemisphere \cite{lee2020perimeter-defense}. 

\begin{figure}[t]
\centering
\includegraphics[width=.48\textwidth]{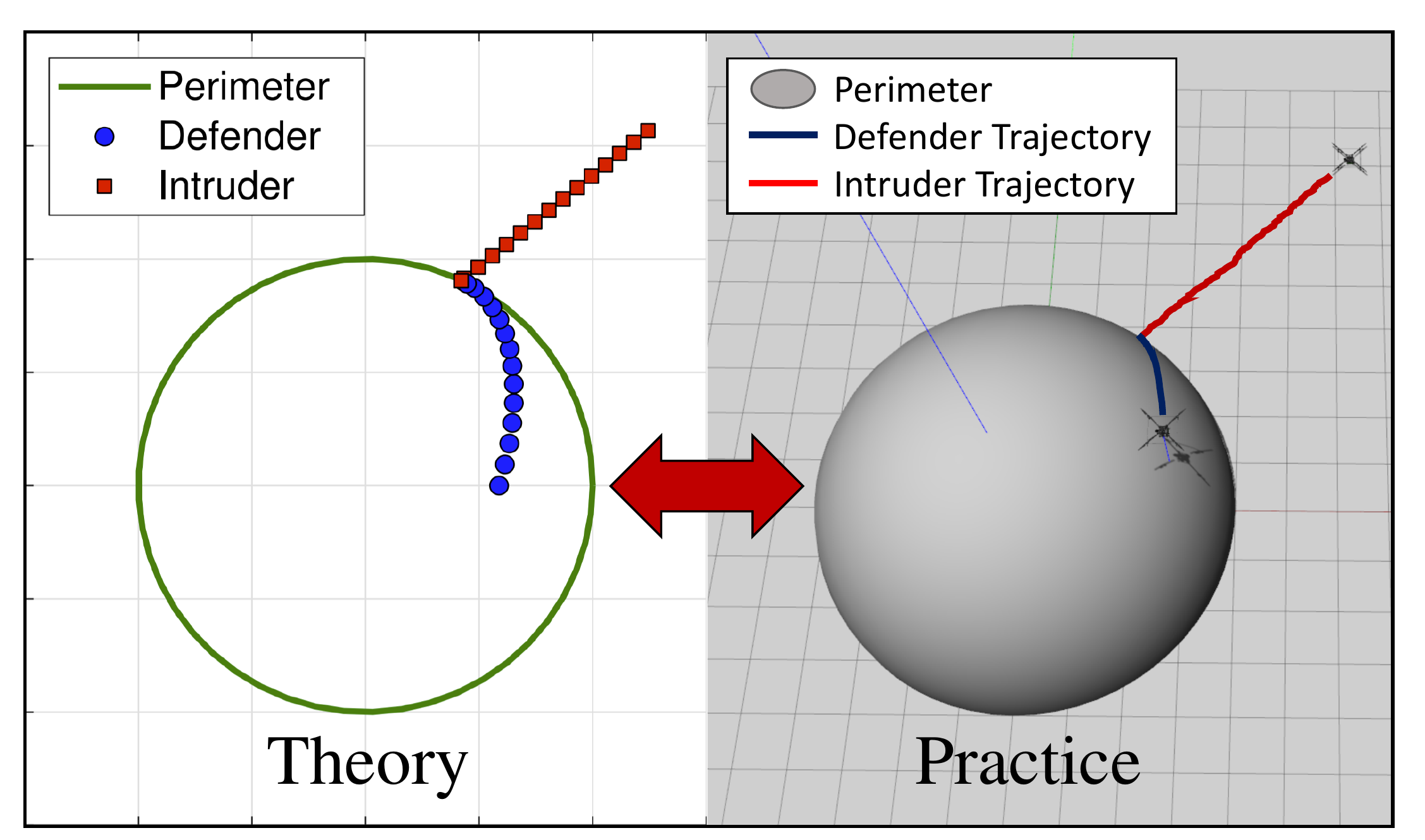}
\caption{
Hemisphere perimeter defense game in theory and practice shows the discrepancy in executing optimal strategies.
}
\label{fig:intro}
\end{figure}

There are many challenges in realizing the perimeter defense system with real robots, and the major challenges lie in coping with discrepancy between first-order assumptions of point particles and dynamics of real robots, as shown in Fig. \ref{fig:intro}. This work simulates the perimeter defense based on Unmanned Aerial Vehicles (UAVs). UAVs are deployed in various space such as power plant \cite{lee2020experimental}, penstock \cite{nguyen2019mavnet}, forest \cite{chen2020sloam}, or disaster sites\cite{lee2016drone}, which are good perimeter defense applications.

This paper extends the previous work \cite{lee2020perimeter-defense} on hemisphere defense to apply the derived theory from point particles to real robots. We observe any performance discrepancy in relaxing first-order assumptions and discuss how to reduce such discrepancy. The contributions of the paper are (i) realizing perimeter defense between aerial defender and ground intruder from theory to practice; and (ii) performing parametric analysis of system scales and comparative study of strategies to evaluate the performance discrepancy in relaxing first-order assumptions.

Section II formulates the problem. Section III addresses the transition from point particle to real robot. To evaluate the discrepancy, two metrics are proposed in Section IV. Section V provides experimental results, and Section VI concludes the paper.



\section{Problem Formulation and Preliminaries \label{sec:problem}}
Consider a hemisphere perimeter $O$ with radius of $R$ as shown in Fig. \ref{fig:coordinates}. The defender $D$ is constrained to move on the hemisphere while the intruder $A$ is constrained to move on the ground plane. The positions of the players in spherical coordinates are: $\mf z_D=[\psi_D,\phi_D,R]$ and $\mf z_A=[\psi_A,0,r]$, where $\psi$ and $\phi$ are the azimuth and elevation angles, which gives the relative position as: $\mf z \triangleq [\psi,\phi,r]$, where $\psi\triangleq \psi_A-\psi_D$ and $\phi\triangleq \phi_D$ (see Fig.~\ref{fig:coordinates}). Without loss of generality, we assume the defender's maximum speed is 1. The intruder is assumed to have a maximum speed $\nu \leq 1$ (otherwise, intruder always has a strategy to win the game). We denote that the velocities of defender and intruder are $v_D$ and $v_A$, respectively. The game ends at time $t_f$ with intruder's win if $r(t_f)=R$ and $|\psi(t_f)| + |\phi_D(t_f)|>0$, whereas it ends with defender's win if $\phi_D(t_f)=\psi(t_f)=0$ and $r(t_f)>R$. 
We call $t_f$ as the \textit{terminal time}.

\subsection{Optimal breaching point for point particle}
\begin{figure}[t]
\centering
\includegraphics[width=.4\textwidth]{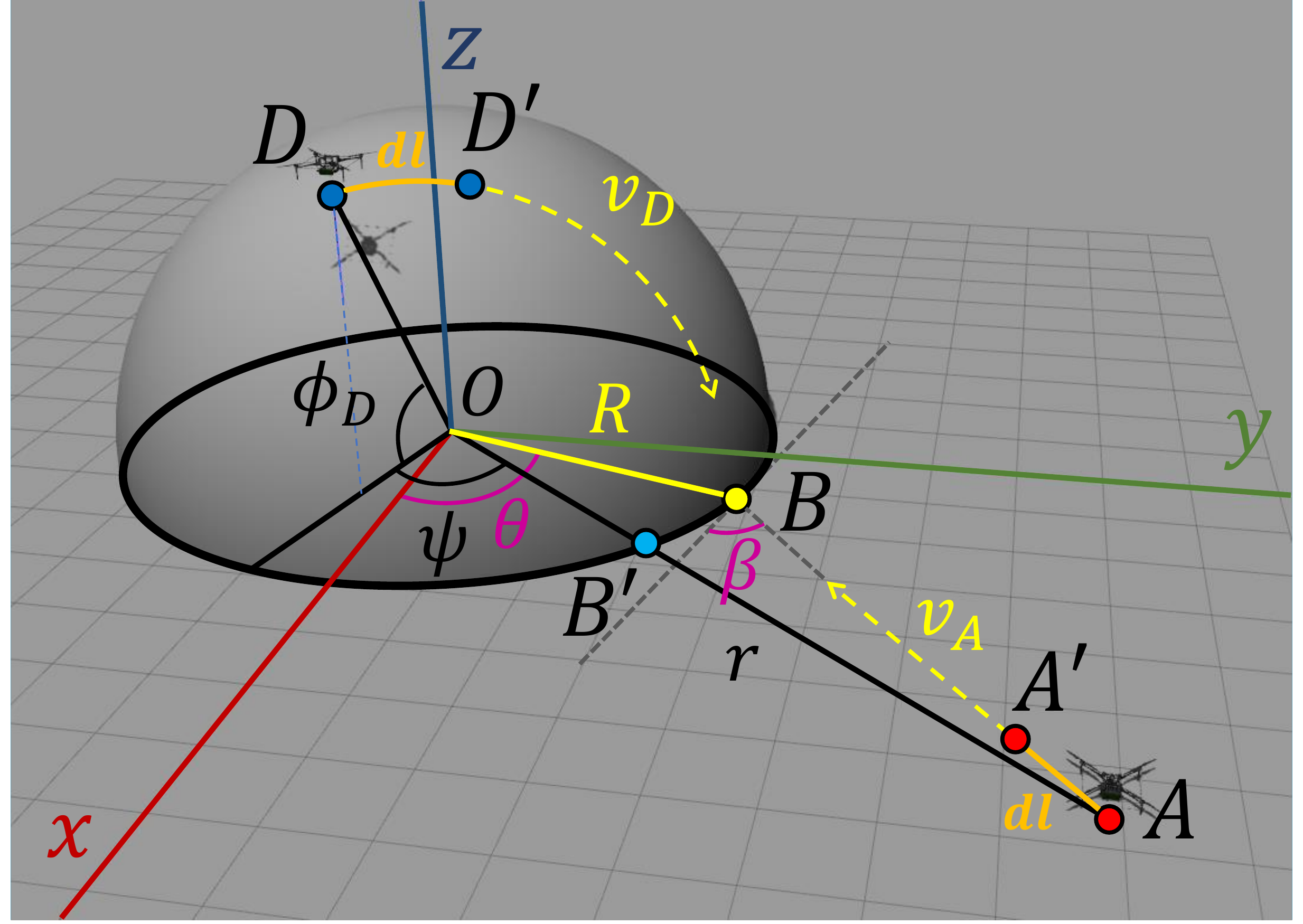}
\caption{The coordinate system and relevant variables.}
\label{fig:coordinates}
\end{figure}

Given $\zd$, $\za$, we call $\textit{breaching point}$ as a point on the perimeter at which the intruder tries to reach the target, as shown $B$ in Fig.~\ref{fig:coordinates}. We call the azimuth angle that forms the breaching point as \textit{breaching angle}, denoted by $\theta$, and call the angle between $(\mf z_A - \mf z_B)$ and the tangent line at $B$ as \textit{approach angle}, denoted by $\beta$. 

It is proved in \cite{lee2020perimeter-defense} that given the current positions of defender $\mf z_D$ and intruder $\mf z_A$ as point particles, there exists a unique breaching point such that the optimal strategy for both defender and intruder is to move towards it, known as \textit{optimal breaching point}. The breaching angle and approach angle corresponding to the optimal breaching point are known as \textit{optimal breaching angle}, denoted by $\theta^*$, and \textit{optimal approach angle}, denoted by $\beta^*$.

As stated in \cite{lee2020perimeter-defense}, although there exists no closed form solution for $\theta^*$ and $\beta^*$, they can be computed at any time by solving two governing equations:

\begin{equation}
\beta^* =  \cos^{-1}\left(\nu\frac{\cos{\phi_D}\sin{\theta^*}}{\sqrt{1-\cos^2{\phi_D}\cos^2{\theta^*}}}\right)
\label{eq:beta}
\end{equation}
and
\begin{equation}
\theta^* = \psi-\beta^*+\cos^{-1}\left(\frac{\cos\beta^*}{r}\right) \label{eq:theta}
\end{equation}

\subsection{Target time and payoff function}
We call the \textit{target time} as the time to reach $B$ and define $\tdd$ as the \textit{defender target time}, $\taa$ as the \textit{intruder target time}, and the following as \textit{payoff} function:
\bql
\pp = \tdd -\taa \label{eq:payoff}
\eql   

The defender reaches $B$ faster if $p<0$ and the intruder reaches $B$ faster if $p>0$. Thus, the defender aims to minimize $p$ while the intruder aims to maximize it. 


\subsection{Optimal strategies and Nash equilibrium  \label{subsec:optimal}}
It is proven in \cite{lee2020perimeter-defense} that the optimal strategies for both defender and intruder are to move towards the optimal breaching point at their maximum speed at any time. 

Let $\Omega$ and $\Gamma$ be the continuous $v_D$ and $v_A$ that lead to $B$ so that 
$\td \triangleq \tdd$ and $\ta \triangleq \taa$, and let $\Omega^*$ and $\Gamma^*$ be the optimal strategies that minimize $\td$ and $\ta$, respectively, then the optimality in the game is given as a Nash equilibrium:
\bql
\pdstar\leq\pdastar\leq\pastar \label{eq:nash}
\eql

\section{From Theory to Practice} \label{sec:from}
This section discusses the transition from theory to practice in executing the optimal strategies for hemisphere perimeter defense game.  

\subsection{From point particle to three-dimensional robot}
The major challenge in bringing the theory closer to practice lies in representing the agents as three-dimensional robot. Previous works in perimeter defense focused on the point particle \cite{liang2019differential,isaacs1999differential,shishika2018local,lee2020perimeter-defense,shishika2020review} to represent the defender and intruder,
and the following are the assumptions made for the point particle, which may not hold true in working with the three-dimensional robot:

\begin{itemize}
  \item It has no volume and thus it is scale invariant
  \item It moves with desired velocity instantly and precisely
  \item If it moves at its maximum speed, the speed is consistent along the trajectory
  \item It can accurately detect other agent's positions and react to it simultaneously 
  \item Optimal trajectory obeys first-order assumptions
\end{itemize}

\begin{figure}[!t]
\centering
\includegraphics[height=2.5cm]{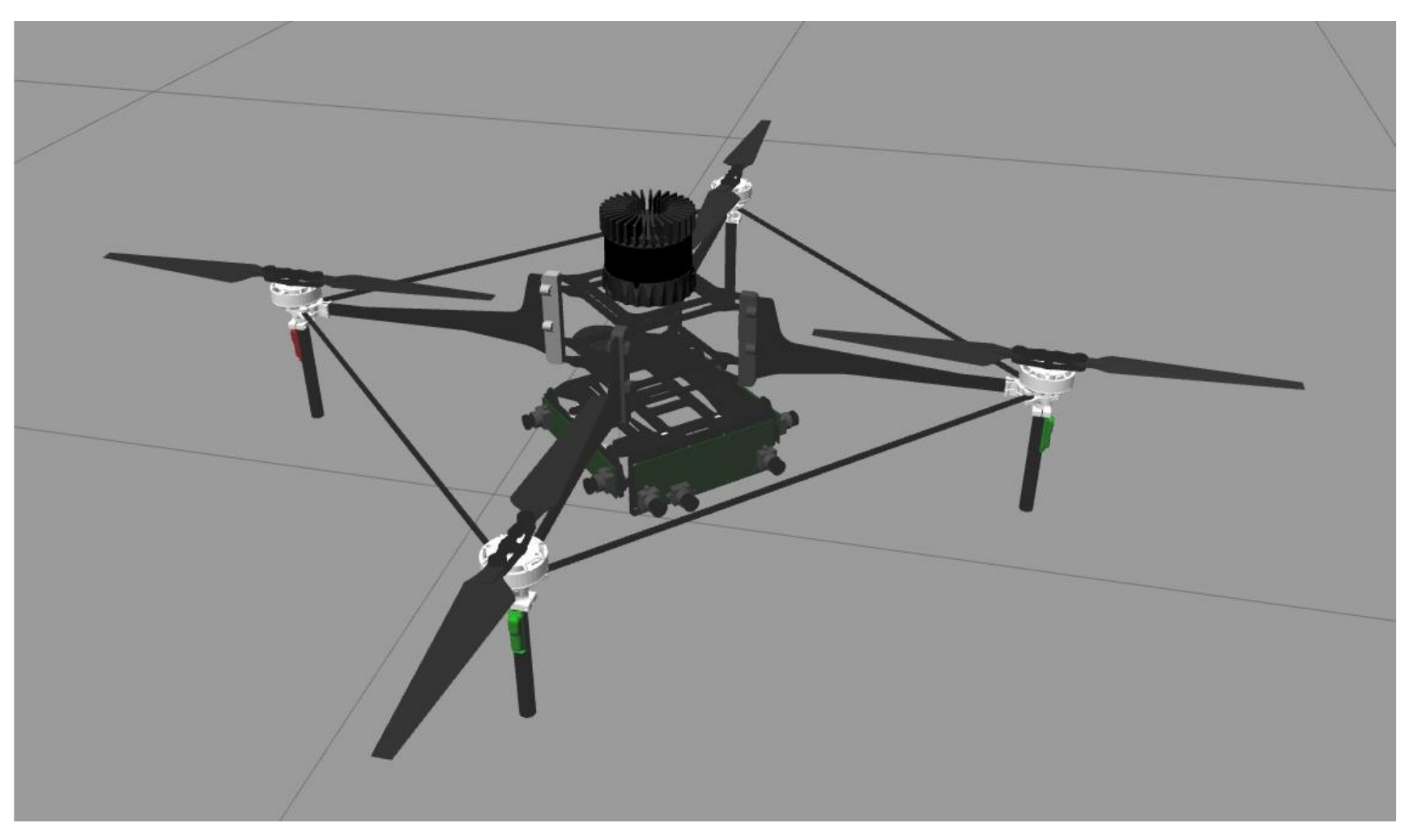}
\caption{
An UAV carrying stereo cameras and Ouster lidar
}
\label{fig:falcon4}
\end{figure}

This work aims to simulate robots using the optimal strategies derived from theory, observe any discrepancy between the performances from theory and practice, and discuss how relaxing aforementioned assumptions would lead to the discrepancy. The employed robot is an UAV in Fig.~\ref{fig:falcon4}. It has a dimension of 735mm $\times$ 735mm $\times$ 200mm and the mass of 1.8kg. This UAV well represents a three-dimensional robot relaxing the point particle assumptions.

\begin{figure}[!b]
\centering
\includegraphics[height=5.6cm]{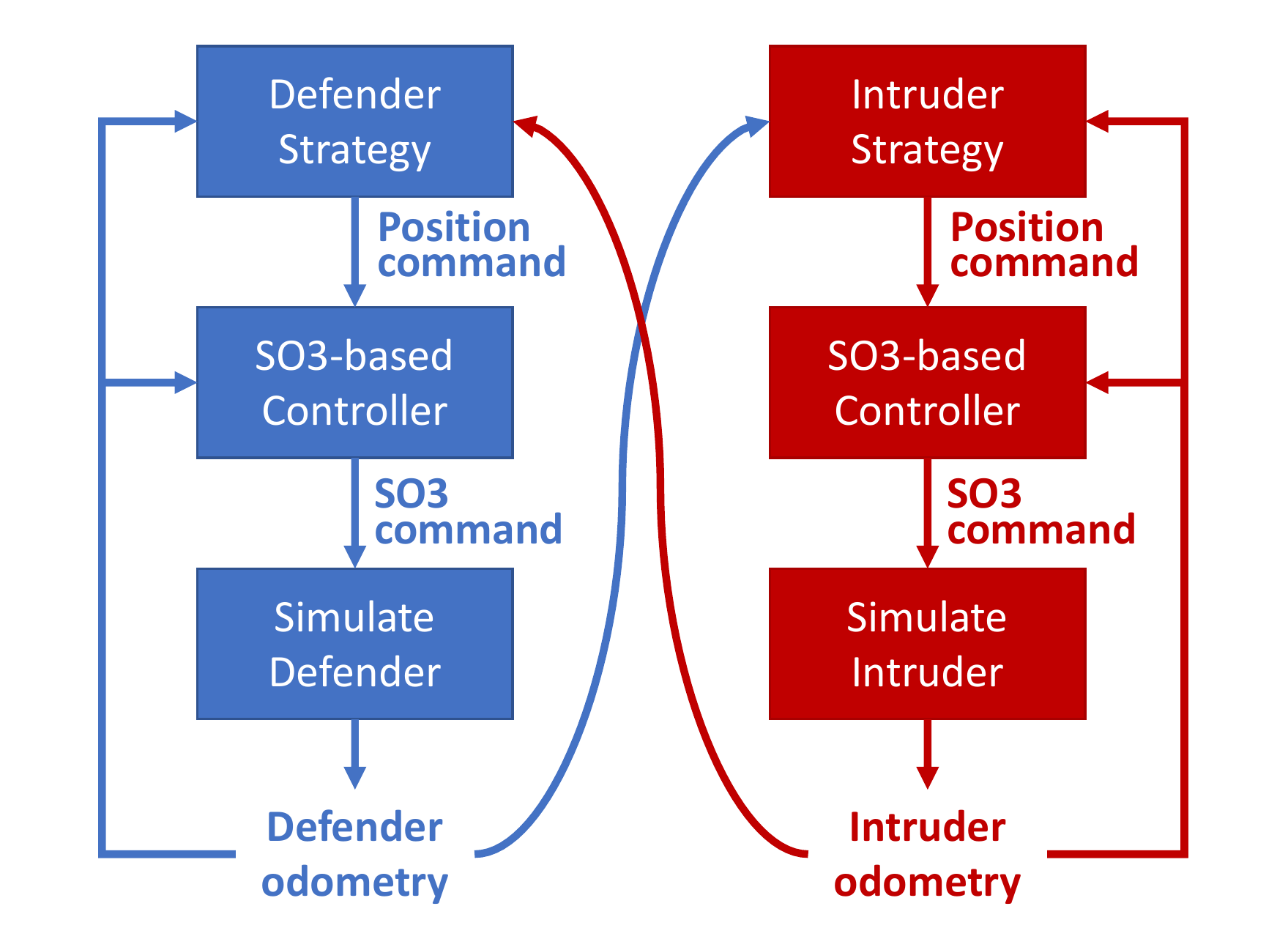}
\caption{
Overall controller for cross feedback system
}
\label{fig:diagram}
\end{figure}

\subsection{Optimal breaching point in practice}
In practice, we relax the assumption of scale invariance. The radius of hemisphere becomes a relevant variable $R$, as denoted in Fig.~\ref{fig:coordinates}. Accordingly, \eqref{eq:theta} in practice becomes

\bql
\theta^* = \psi-\beta^*+\cos^{-1}\left(\frac{R\cos\beta^*}{r}\right) \label{eq:theta2}
\eql
although \eqref{eq:beta} stays the same in practice since all the parameters are scale invariant. We can solve for $\beta^*$ and $\theta^*$ by solving two governing equations \eqref{eq:beta} and \eqref{eq:theta2}, and obtain the corresponding optimal breaching point in practice.

\subsection{Controller}
As mentioned in the previous assumptions \cite{lee2020perimeter-defense}, the controller used for theoretical approach was simply perfect. Both defender and intruder reacted to each other without any delay, and desired dynamics were enforced precisely and instantly to result in continuous optimal trajectories as shown in Fig. \ref{fig:intro}. The agents also identify each other's states without any error at any time (e.g. perfect state estimation). 

The overall controller for practical approach is illustrated in Fig. \ref{fig:diagram}. All modules represent the nodes in ROS and are connected by rostopics for real-time implementation. For each agent, the strategy module subscribes to the ground truth odometry that are published by both UAVs to calculate position command. In this way, state estimation is based on ground truth and becomes highly accurate. Then, the position command is fed into the SO3-based controller based on \cite{mellinger2011minimum} to generate SO3 command. The SO3 commands consist of thrust and moment that control the robot in low level. Finally, the simulated robot publishes the ground truth odometry to be fed into the SO3-based controller for closed loop control. Note that the defender and intruder strategy modules take in both agents' odometry. 
For this cross feedback system, the two odometry data are stored into queues and used for synchronized computation of the optimal strategy.

\subsection{Executing optimal strategies}
As stated in Sec.\ref{subsec:optimal}, both defender and intruder move at their maximum speeds in executing optimal strategies in theory. It is also worth noting that the optimal trajectories are continuously changing based on the configuration of agents at given time because the optimal breaching point is newly calculated at any time to reflect the change in positions of the defender and intruder. 

For this reactive and adversarial environment, we enforce the players to follow an infinitesimal segment of computed optimal trajectory at any time to maintain consistent maximum speeds. In Fig. \ref{fig:coordinates}, given the optimal breaching point $B$ and current configurations of defender and intruder $\mf z_D=[\psi_D,\phi_D,R]$ and $\mf z_A[\psi_A,0,r]$, their current optimal trajectories are $\widearc{DB}$ and $\overline{AB}$, respectively. Based on the movements of defender and intruder, the optimal trajectories as well as the optimal breaching point may change over time, so we set the goal trajectory $\widearc{DD'}$ and $\overline{AA'}$ as segments of original trajectories by taking the length of $dl$ and $dl'$ for defender and intruder, respectively. We give position command towards the end point of the infinitesimal segment, which can be viewed as controlling an instantaneous velocity of agents. 

First, we aim to compute the position of $D'[x_{D}, y_{D}, z_{D}]$ after infinitesimal defender movement $dl$. For simplicity, assume $D$ is on the zx-plane (i.e. $\psi_D = 0$) by rotating the hemisphere by $-\psi_D$ about the z-axis, and we will rotate the coordinates by $\psi_D$ back to the original orientation at the end. Given the radius $R$, azimuth angle $\phi_D$, and $dl$, we know the following: (i) $D$ is on the hemisphere $O$ with a center on the origin and a radius of $R$; (ii) The arc distance between $D$ and $D'$ is $dl$; (iii) $D'$ is on the plane $OBD$. The conditions (i) and (ii) give
\bql
x_D^2+y_D^2+z_D^2 = R^2 \text{\; and} \label{eq:con1}
\eql
\bql
(x_D-R\cos\phi_D)^2+y_D^2+(z_D-R\sin\phi_D)^2 = dl^2 \label{eq:con2}
\eql
We know the equation of a plane is given by
\bql
ax+by+cz=d \label{eq:plane}
\eql
Therefore, the plane $OBD$ is uniquely determined by three points $O(0,0,0)$, $B(R\cos\theta, R\sin\theta, 0)$, and $D(R\cos\phi_D$, 0, $R\sin\phi_D)$. With the condition (iii), \eqref{eq:plane} becomes
\bql
(\sin\phi_D\sin\theta) x_D-(\sin\phi_D\cos\theta)y_D-(\cos\phi_D\sin\theta)z_D =0 \label{eq:con3}
\eql
Together with \eqref{eq:con1}, \eqref{eq:con2} and \eqref{eq:con3}, we get
\bnl
x_D =& R\cos\phi_D - \frac{dl^2\cos\phi_D}{2R} + \frac{dl\sin^2\phi_D\cos\theta}{2R}\cdot T \notag\\ 
y_D =& \frac{dl\sin\theta}{2R}\cdot T \notag\\ 
z_D =& R\sin\phi_D - \frac{dl^2\sin\phi_D}{2R} -\frac{dl\sin\phi_D\cos\theta\cos\phi_D}{2R}\cdot T
\label{eq:xd}
\enl
where
\beq
T = \sqrt{\frac{4R^2-dl^2}{1-\cos^2\phi_D\cos^2\theta}}
\eeq

The final computation of the position $x_D,y_D,z_D$ is summarized in Algorithm \ref{alg1}. Notice that line 4 and 5 rotate the system back to the original configuration. Similarly, the position of $A'[x_A,y_A,z_A]$ after infinitesimal intruder movement $dl'$ is computed in Algorithm \ref{alg2}, and overall agents' strategies are summarized in Algorithm \ref{alg3}. 

\begin{algorithm}[!b]
\caption{[$x_D, y_D, z_D$] = MoveDefender($\psi_D,\phi_D,\theta^*$)}
\label{alg1}
\begin{algorithmic}[1]
\REQUIRE $\psi_D,\phi_D,\theta^*$ 
\ENSURE Updated defender positions $x_D, y_D, z_D$
\STATE Set defender's small movement $dl$
\STATE $\theta=\theta^*-\psi_D$
\STATE Evaluate $x_{D},y_{D},z_{D}$ using \eqref{eq:xd}
\STATE $x_D = x_{D}\cos\psi_D - y_{D}\sin\psi_D$
\STATE $y_D = x_{D}\sin\psi_D + y_{D}\cos\psi_D$
\end{algorithmic}
\end{algorithm}

\begin{algorithm}[!b]
\caption{[$x_A, y_A, z_A$] = MoveIntruder($x_A, y_A, z_A, \theta^*$)}
\label{alg2}
\begin{algorithmic}[1]
\REQUIRE $x_A, y_A, z_A, \theta^*$
\ENSURE Updated intruder positions $x_A, y_A, z_A$
\STATE Set intruder's small movement $dl'$
\STATE $x_B = R\cos\theta^*$
\STATE $y_B = R\sin\theta^*$
\STATE $l = \sqrt{(x_B-x_A)^2+(y_B-y_A)^2}$
\STATE $x_A = x_A + (x_B-x_A)dl'/l$
\STATE $y_A = y_A + (y_B-y_A)dl'/l$
\end{algorithmic}
\end{algorithm}

\begin{algorithm}[!b]
\caption{Defender/Intruder Strategy}
\label{alg3}
\begin{algorithmic}[1]
\REQUIRE Agent odometry $x_D, y_D, z_D, x_A, y_A, z_A$
\ENSURE Position command (updated $x_D, y_D, z_D$ for defender and updated $x_A, y_A, z_A$ for intruder)
\STATE Initialize $r, \psi_A, \psi_D, \phi_D$ using odometry
\IF{$\phi_D>0$ or $|\psi_A-\psi_D|>0$}
\STATE Compute optimal breaching angle $\theta^*$
\IF{Defender Strategy}
\STATE{[$x_D, y_D, z_D$] = MoveDefender($\psi_D,\phi_D,\theta^*$)}
\ELSIF{Intruder Strategy}
\STATE{[$x_A, y_A, z_A$] = MoveIntruder($x_A, y_A, z_A, \theta^*$)}
\ENDIF
\ENDIF
\end{algorithmic}
\end{algorithm}

\section{Discrepancy Analysis \label{sec:performance}}
In this section, we propose two metrics to evaluate the discrepancy between the performances from theory and practice. One metric measures the discrepancy from relaxing the scale invariant assumption made for point particle. This metric is crucial since the discrepancy between small and large scales would not allow scalability, which is essential for real-world applications. The other metric investigates if the traversed trajectory is the same as the optimal trajectory derived from theory by comparing its performances with the performance of other baseline strategy. We chose these metrics to evaluate the discrepancy because other point particle assumptions (e.g. instant velocity change, simultaneous detection and reaction, and no fluctuation in maximum speed) highly depend on the hardware. A robot with better computing power will be closer to meet these assumptions, so we focus on the discrepancy in system scales and strategies.

\subsection{Discrepancy in system scales}
\begin{figure}[!b]
\centering
\includegraphics[width=6.9cm]{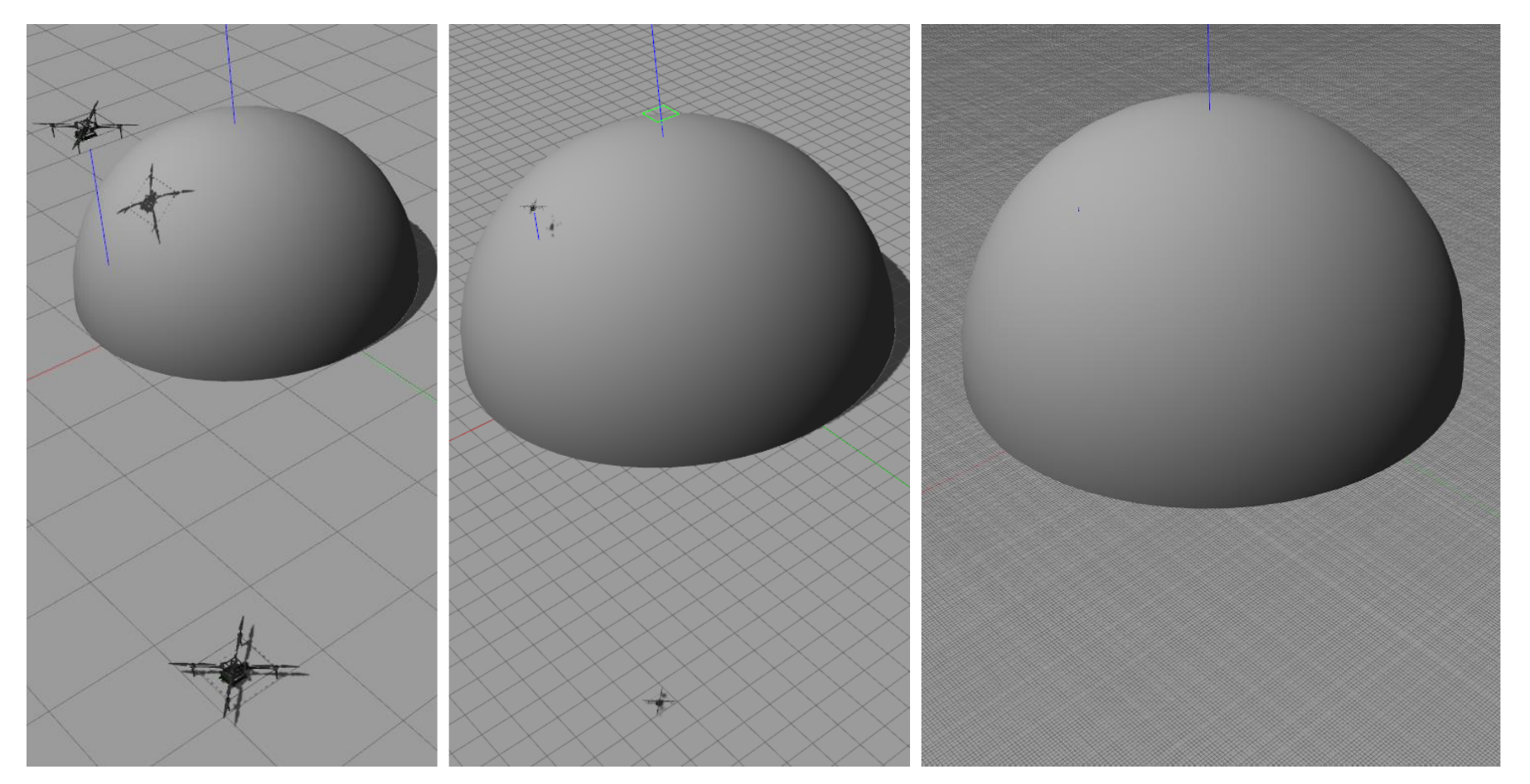}
\caption{
Different system scales of environment and UAVs with a fixed scale. \textbf{Left}: Radius of 3. \textbf{Middle}: Radius of 10. \textbf{Right}: Radius of 100. 
}
\label{fig:3hemi}
\end{figure}

Fig. \ref{fig:3hemi} shows different scales of environment (e.g. hemisphere) for perimeter defense game. Since the robot size is fixed, this variance may impact the dynamics of the UAV. For instance, if the robot traverses along the surface of small hemisphere, it may enforce the controller to execute abrupt turns and the resulting trajectory may not exactly follow the hemisphere surface. 

\begin{figure}[!b]
\centering
\includegraphics[width=5.3cm]{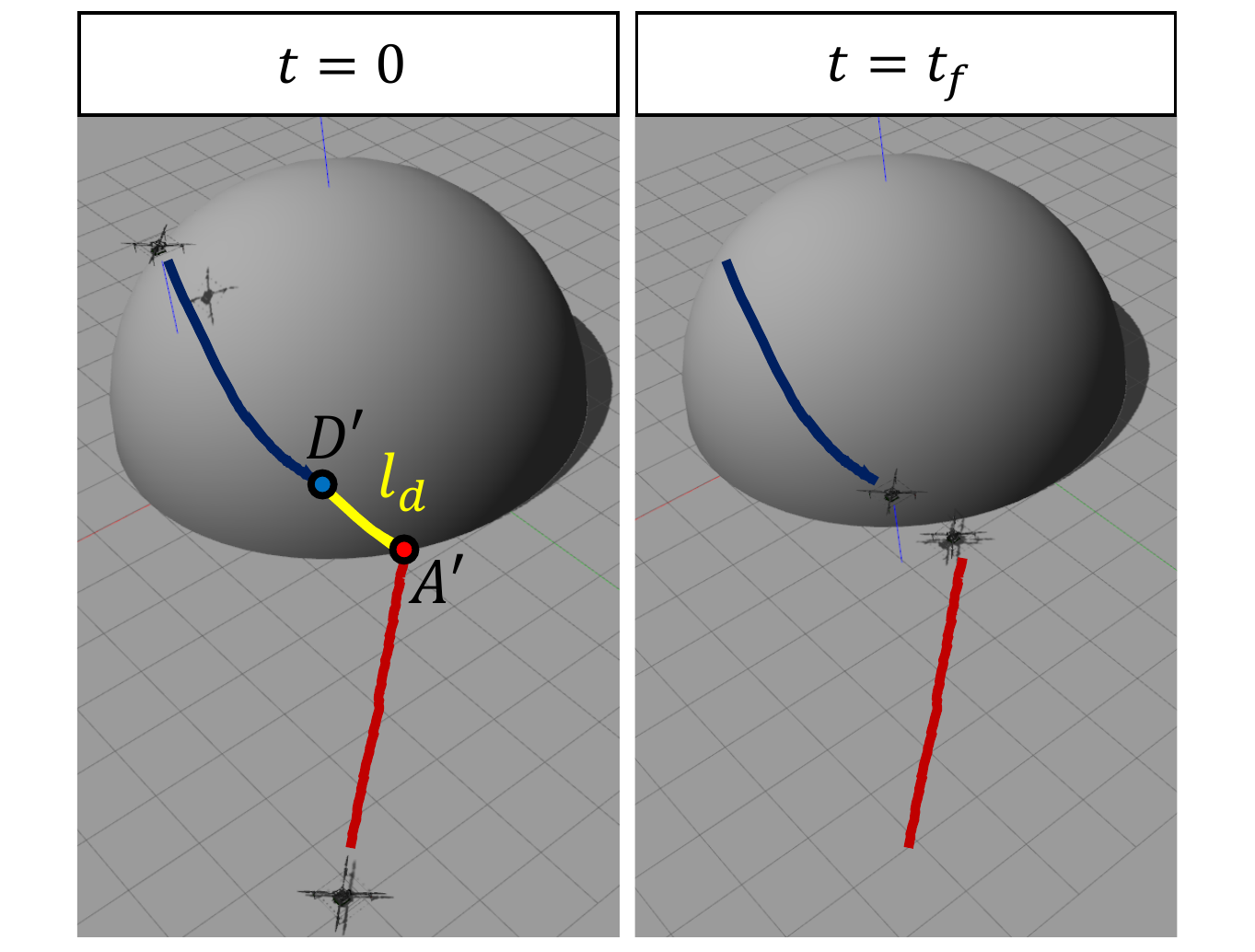}
\caption{
Game setting for computing $l_d$. \textbf{Left}: Initial configuration when the game begins. \textbf{Right}: Final configuration at terminal time $t_f$.
}
\label{fig:scale}
\end{figure}

To account for such performance discrepancy, we propose a metric $l_d$ as shown in Fig. \ref{fig:scale}. This metric computes the geodesic between $D'$ and $A'$, which represent the positions of defender and intruder at terminal time $t_f$, respectively. We purposely set the game for the intruder to win (e.g. intruder reaches to perimeter before defender does) because more interesting dynamics are happening for defender so we fix the behavior of intruder and evaluate $l_d$ that is normalized by radius for different set of system scales. We also compare $l_d$ with the outcome derived from first-order assumptions and observe how the relaxed point particle assumptions impact $l_d$, which would be closer to the theoretical outcome in ideal conditions.

\subsection{Discrepancy in strategies}
\begin{figure}[t]
\centering
\includegraphics[width=5.3cm]{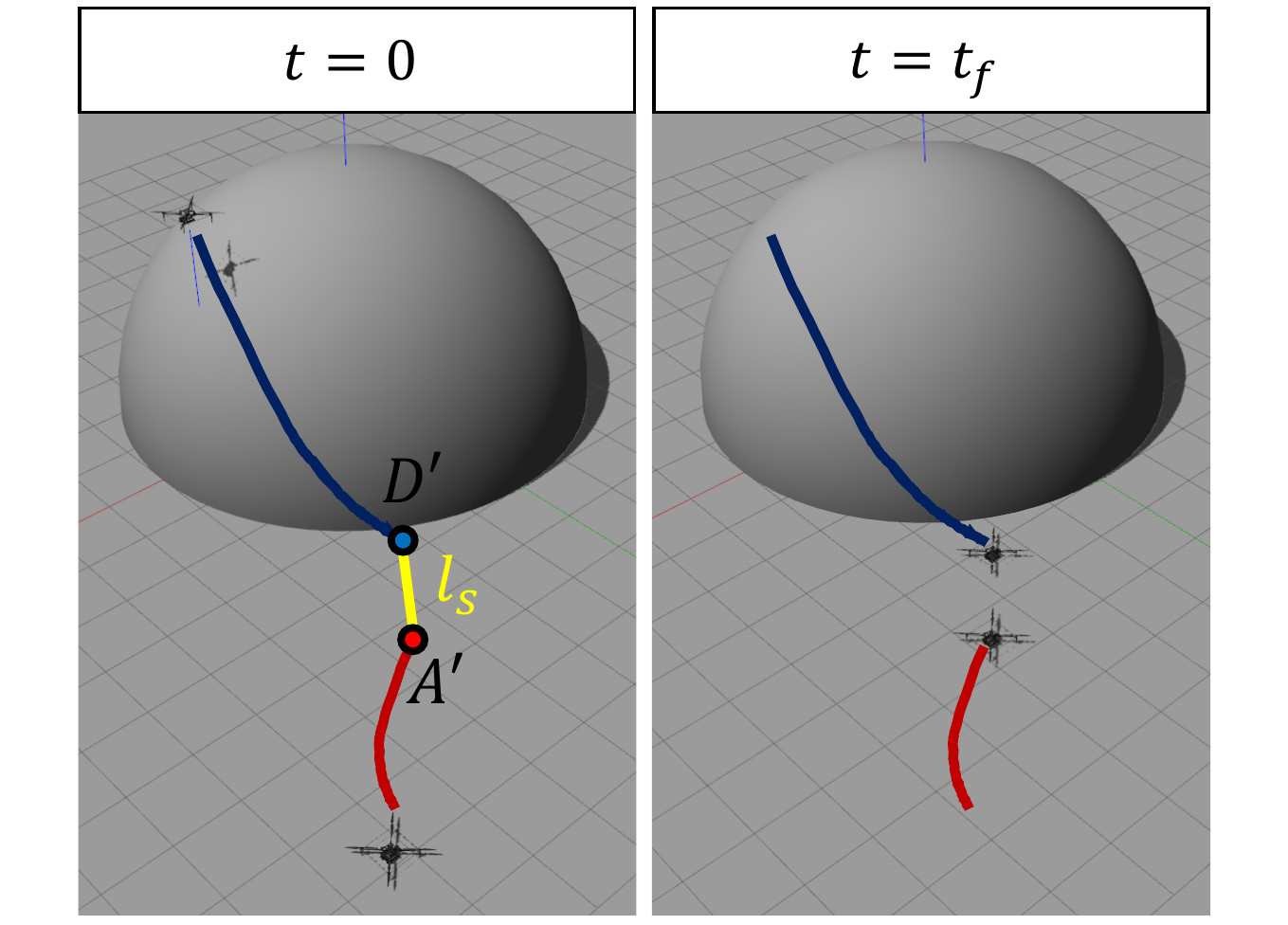}
\caption{
Game setting for computing $l_s$. \textbf{Left}: Initial configuration when the game begins. \textbf{Right}: Final configuration at terminal time $t_f$.
}
\label{fig:strategy}
\end{figure}

We define a metric $l_s$, shown as $l_s$ in Fig. \ref{fig:strategy}, to understand the discrepancy in different defender strategies. This metric measures the distance between $D'$ and $A'$ at terminal time $t_f$, which represents the distance between the agents at the end of the game. To obtain this metric, we intentionally set the game for the defender to win so that $D'$ lies on the base of the hemisphere. In this manner, any plausible defender strategy can be evaluated because the game ends only if the defender finishes executing its strategy. If $l_d$ is to be used as the metric, the game would end before defender fully executes the strategy, which makes it challenging to compare the performance.

In comparing strategies, we run two defender strategies: (1) optimal strategy that moves defender towards the breaching point $B$ at its maximum speed at any time; and (2) baseline strategy that moves defender at its maximum speed at any time towards the azimuth angle of intruder, which is towards $B'$ that is an intersection of $\overline{OA}$ and the base of hemisphere as shown in Fig. \ref{fig:coordinates}. A strategy with larger $l_s$ outperforms the other since it secures more time to defend the intruder.

\section{Experiments \label{sec:experiments}}
This section evaluates the performance discrepancy between theory and practice based on system scales and strategies. We run all experiments using Gazebo with ROS.

\subsection{Parametric analysis of system scales}
In the parametric analysis, we vary the system scales by changing the radius of the hemisphere. A total of 22 different radii from 3 to 300 are used, and multiple experiments are run for each radius to compute $l_d$. Given radius $R$, the experiments are run with an initial configuration $\mf z=[\psi,\phi,r]=[0.9, 0.3\pi, 2R]$, infinitesimal movement $dl=dl'=0.72$ (lead to $v_D=v_A=0.8$), and $\nu=1$.

\begin{figure}[!b]
\centering
\includegraphics[width=7.678cm]{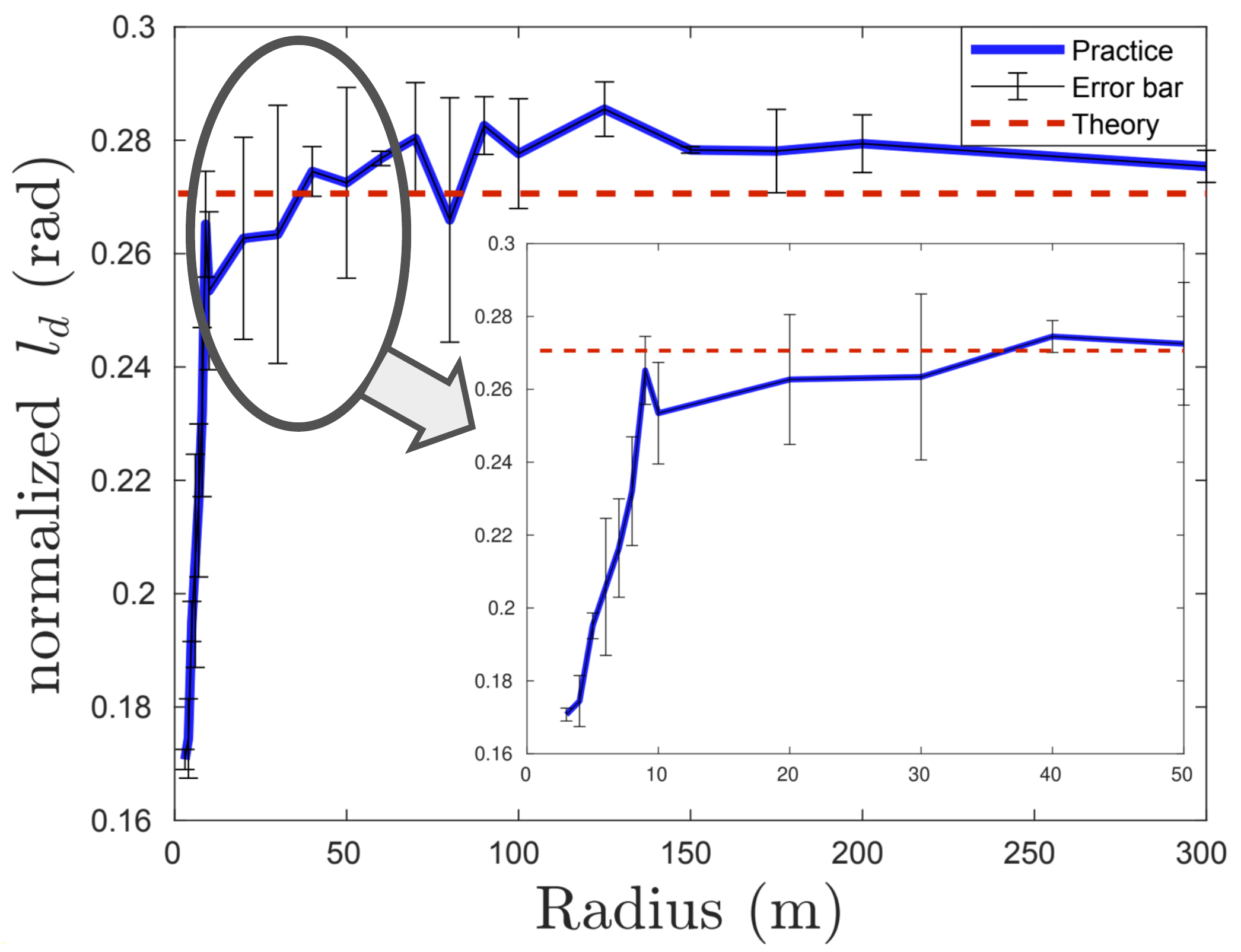}
\caption{Parametric analysis. Magnified view shows small radius regime.}
\label{fig:analysis}
\end{figure}

\begin{figure*}[!t]
\centering
\includegraphics[height=3.31cm]{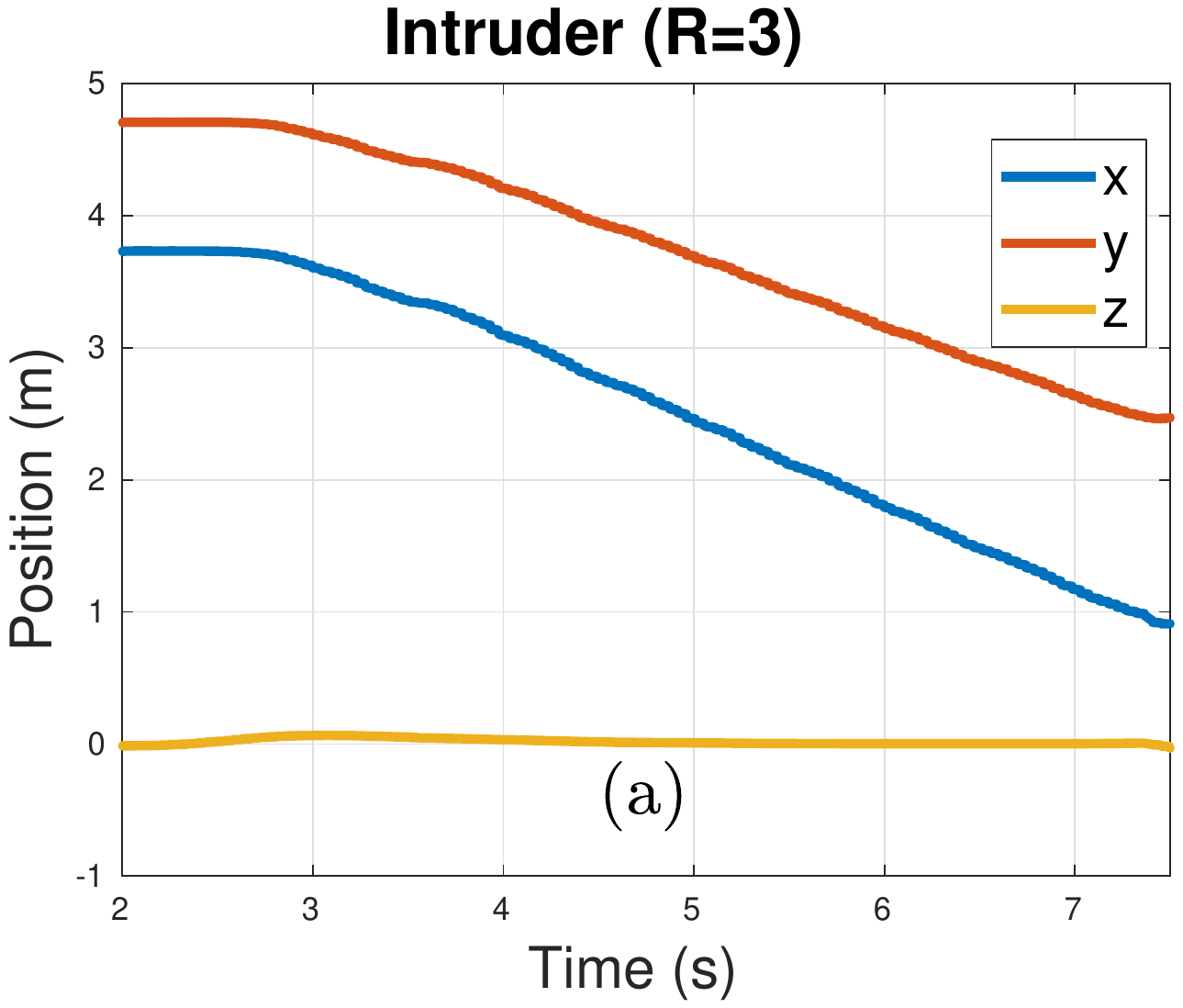}
\includegraphics[height=3.31cm]{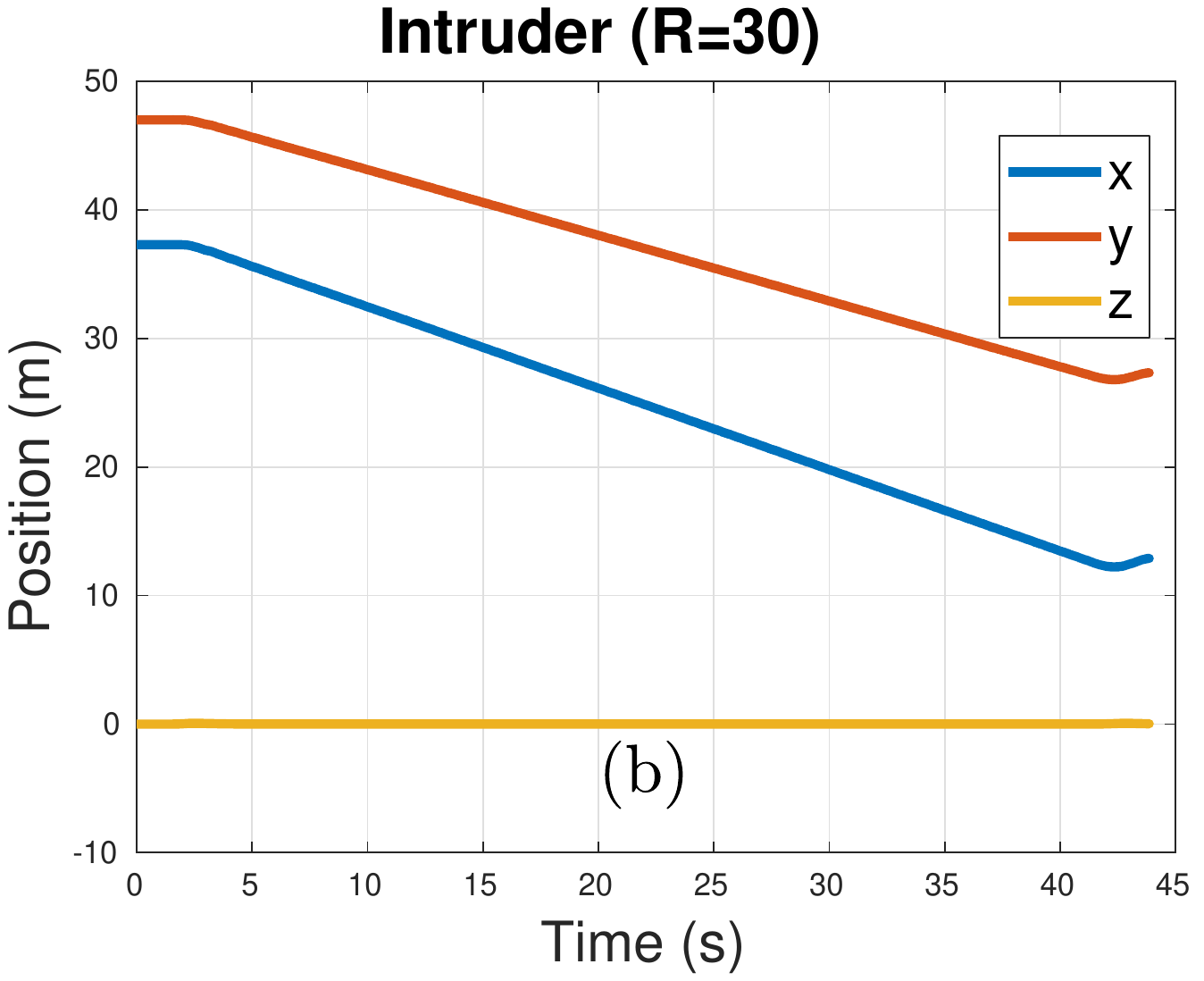}
\includegraphics[height=3.31cm]{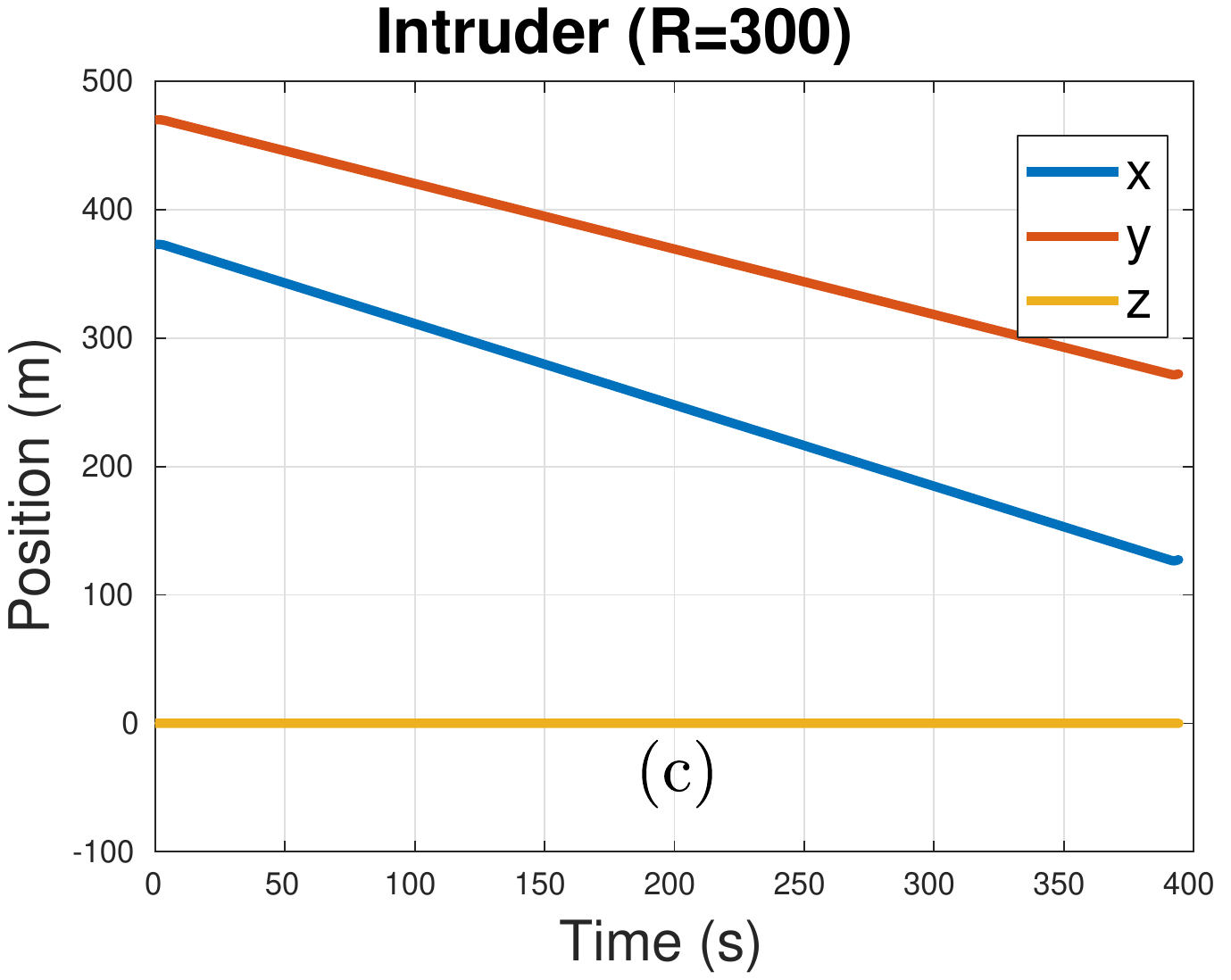}\\
\includegraphics[height=3.31cm]{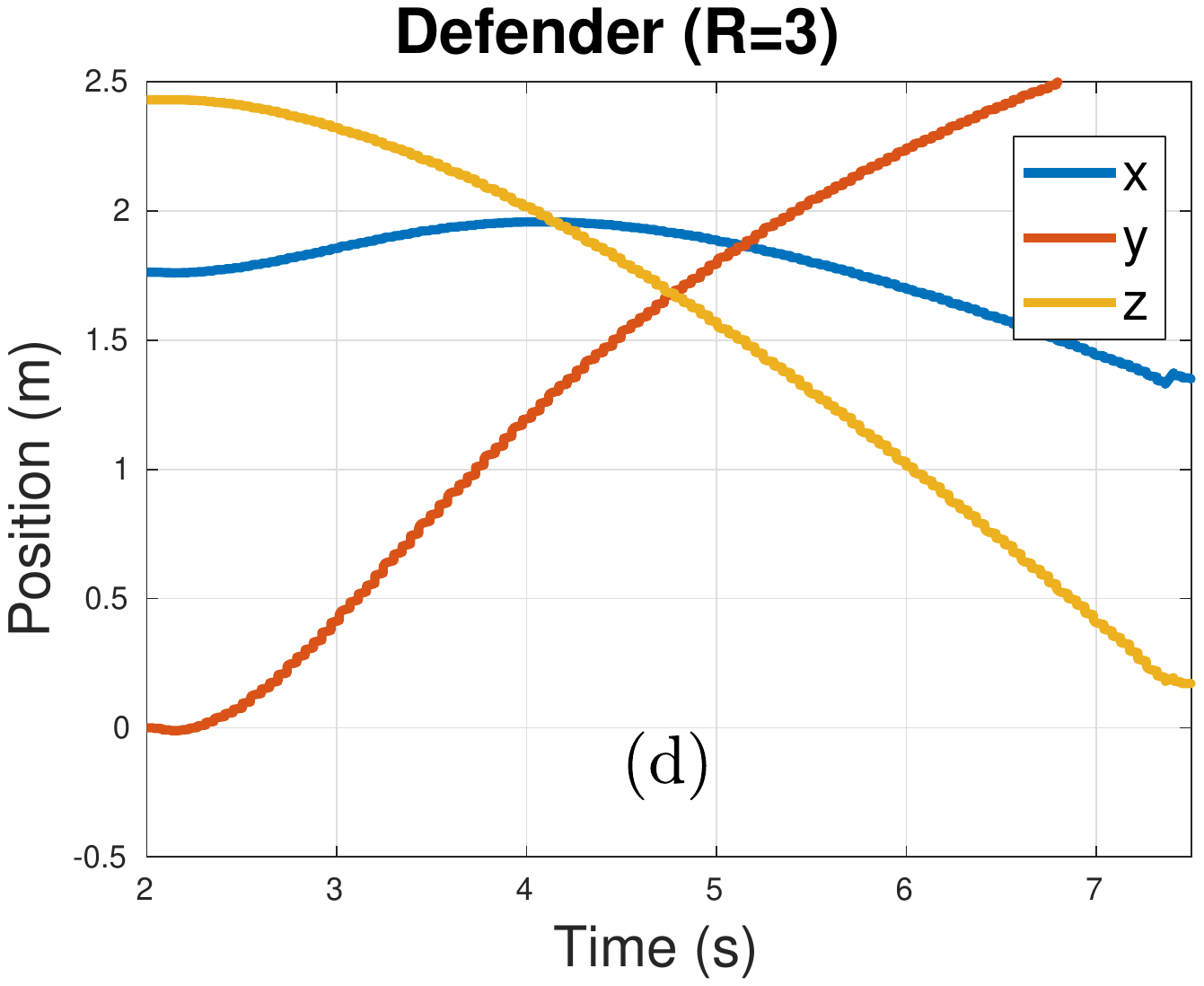}
\includegraphics[height=3.31cm]{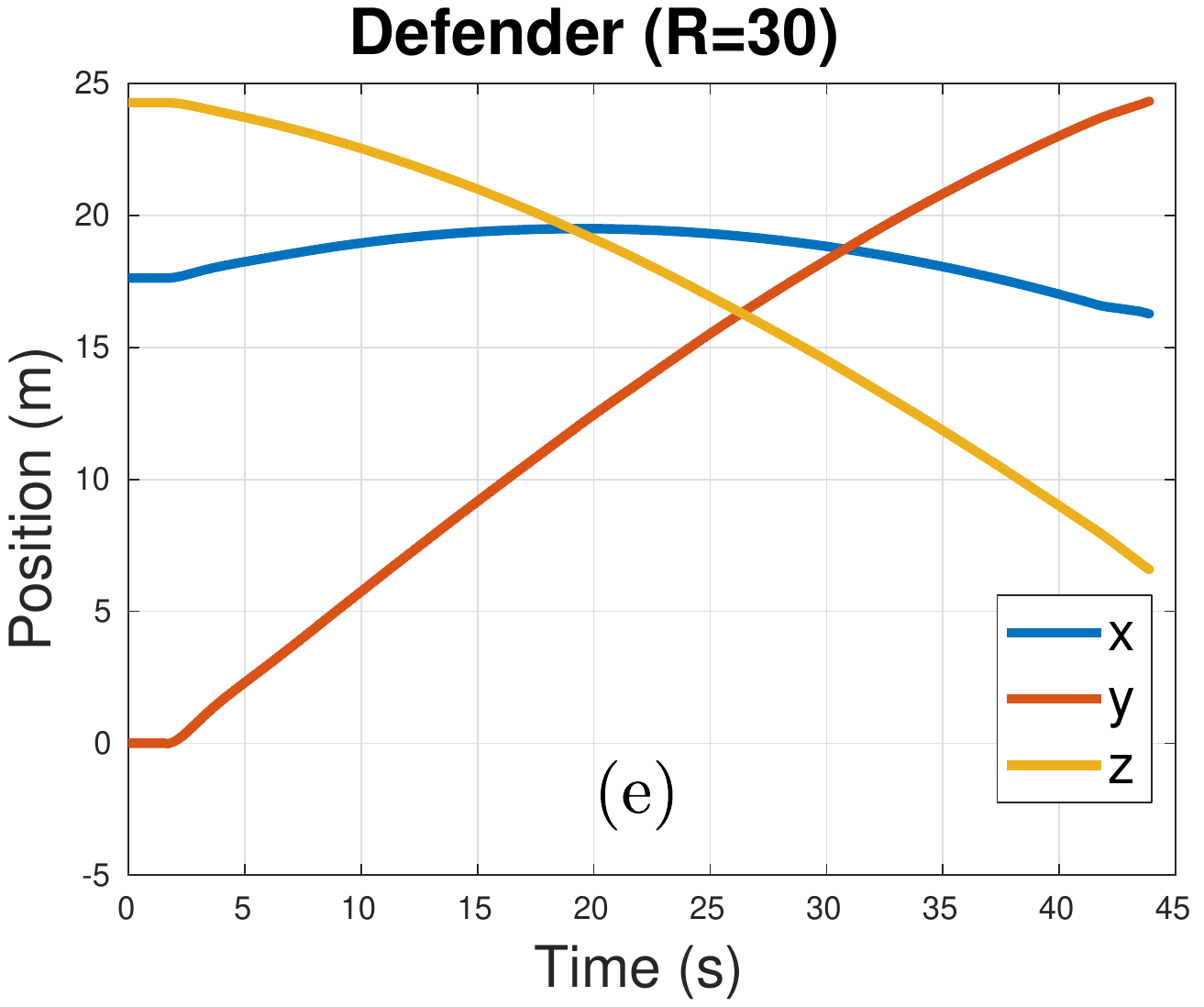}
\includegraphics[height=3.31cm]{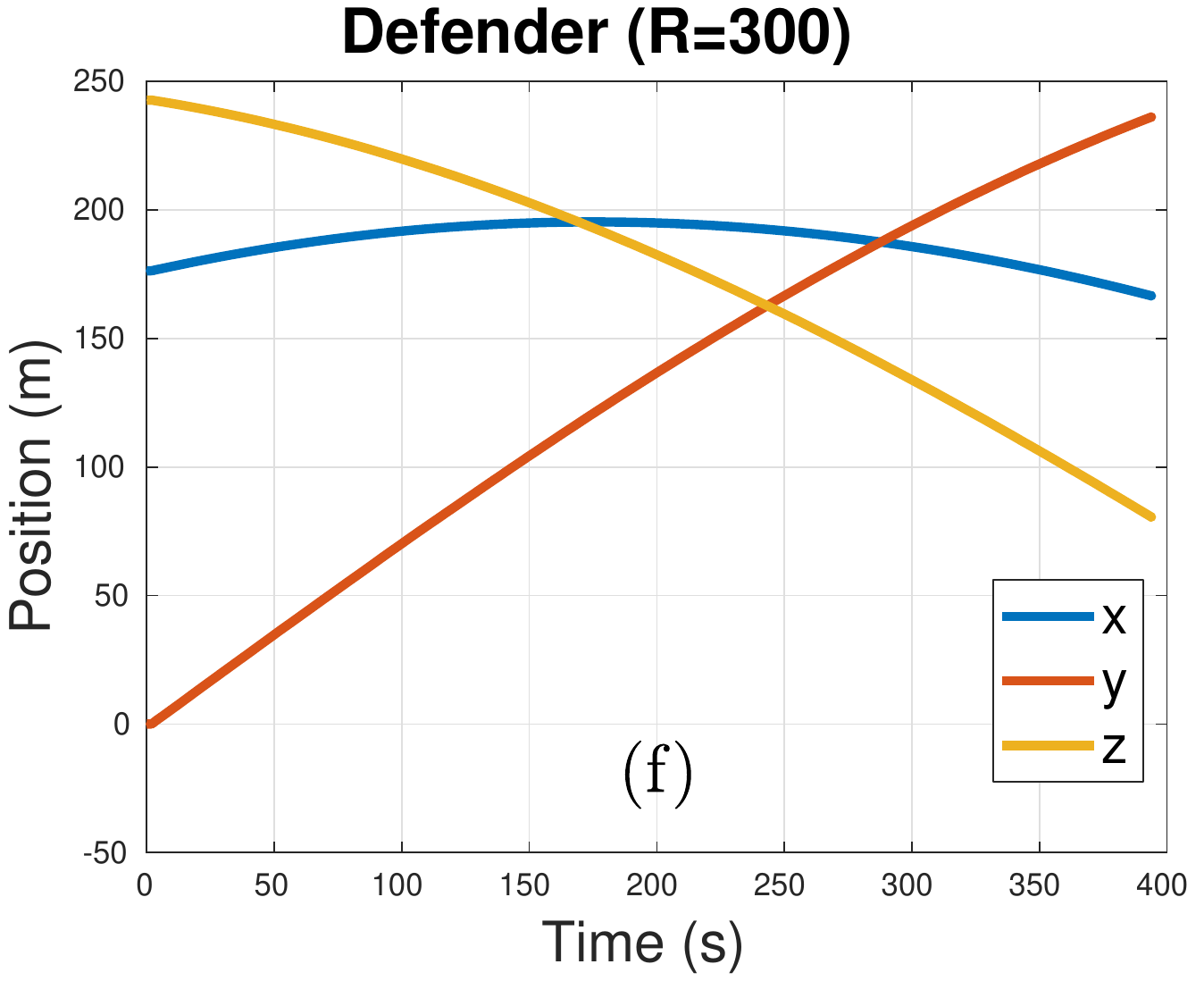}
\caption{
The xyz positions of the defender and intruder with three set of radii. UAV trajectories converge to first-order assumptions as $R$ increases.}
\label{fig:traj}
\end{figure*}

\begin{figure}[!t]
\centering
\includegraphics[width=5.78cm]{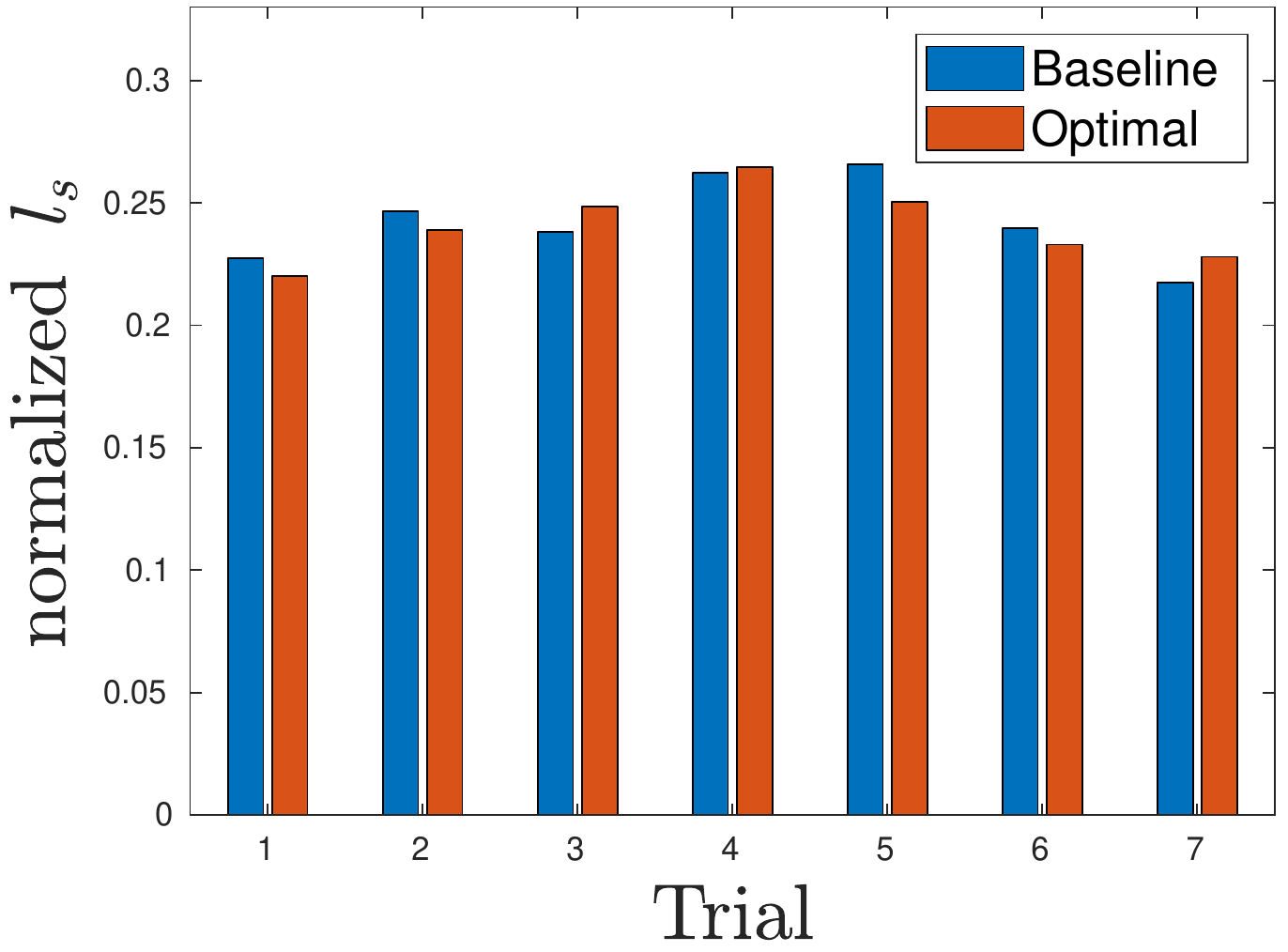}
\caption{
Comparative study of baseline and optimal strategy.}
\label{fig:comparative}
\end{figure}

Fig. \ref{fig:analysis} shows the change in $l_d$ as a function of radius. $l_d$ is normalized by radius so it should be consistent in theory regardless of system scales. It can be seen that $l_d$ stays relatively small in the regime of small radius (e.g. $3\leq R \leq50$) and converges to the value derived from first-order assumptions as $R$ increases. The error bar indicates that the standard deviation decreases as $R$ increases. Fig. \ref{fig:traj} confirms that small radius regime may result in non-smooth trajectories and the UAV trajectories converge to first-order assumptions as $R$ increases. We infer that in the small radius regime, UAV dynamics is sensitive to control inputs since the scale of robot is comparable to the scale of environment. The UAV may accelerate well to get up to the maximum speed so $l_d$ is lower and inconsistent with the outcome from the first-order assumptions.

In this way, the point particle assumption that agents move with desired velocity instantly is much relaxed. Accordingly, to reduce the discrepancy between theory and practice, a larger scale of environment relative to that of robot is preferred, and sensitivity analysis for the system would help the robot better obey the scale invariant assumptions.

\subsection{Comparative study of strategies}
For this study, we compute $l_s$ for optimal and baseline strategies. Seven trials with different configuration are conducted to observe the performance discrepancy. An infinitesimal defender movement $dl=1.36$ (lead to $v_D=1.0$), an infinitesimal intruder movement $dl'=0.36$ (lead to $v_A=0.6$), and $\nu=1$ are used to guarantee that the defender reaches to the base of hemisphere earlier than the intruder does.

Fig. \ref{fig:comparative} shows the variations of $l_s$ normalized by radius for the two strategies in different trials. Although the discrepancy is very small, it is worth noting that the baseline strategy sometimes outperforms the optimal strategy. This tendency supports the claim that there exists discrepancy between theory and practice and that the optimal strategy derived from first-order assumptions could perform worse than other strategies in high-order dynamics. This discrepancy is due to the delay in executing the optimal strategy because computing the optimal breaching point takes some time. As a reference, one cycle of the cross feedback system in Fig. \ref{fig:diagram} takes about 70 msec, and the optimal breaching point calculation takes the major portion of it, which is about 10 msec. To compensate for the delay, robots with higher computing power can be employed.

\section{Conclusion \label{sec:conclusion}}
This paper aims to apply the theory with first-order assumptions derived from the hemisphere perimeter defense game to robots with realistic models, and observe performance discrepancy in relaxing point particle assumptions. We study the transition from theoretical point particles to practical three-dimensional robots using UAV models in Gazebo. To evaluate the performance discrepancy, two analyses are conducted. In the parametric analysis, it is found that UAV trajectories converge to first-order assumptions as the radius of the hemisphere increases. The comparative study shows that the optimal strategy derived from first-order assumptions can perform worse than other strategies in high-order dynamics due to the delay in executing the strategy. To cope with this discrepancy, the future work will focus on changing the first-order model to better capture the high dynamics of the UAVs and design pursuit strategies for the new model.

\bibliography{main}
\bibliographystyle{unsrt}

\end{document}